\documentclass[journal]{IEEEtran}

\ifCLASSINFOpdf
\else
\fi
\usepackage{cite}

\usepackage{blindtext}
\usepackage{dblfloatfix} 

\usepackage{graphicx}
\usepackage{bm}
\usepackage{amssymb}
\usepackage{upgreek}
\usepackage[normalem]{ulem}

\makeatletter
\let\MYcaption\@makecaption
\makeatother
\usepackage[font=footnotesize]{subcaption}
\makeatletter
\let\@makecaption\MYcaption
\makeatother

\makeatletter
\def\endthebibliography{%
  \def\@noitemerr{\@latex@warning{Empty `thebibliography' environment}}%
  \endlist
}
\makeatother

\usepackage{amsfonts,amssymb,amsmath, mathrsfs}

\usepackage{xcolor}
\usepackage{amssymb}
\usepackage{multirow}
\usepackage{graphicx}
\usepackage{tabularx}
\usepackage[acronym]{glossaries}
\usepackage{makecell}
\usepackage{rotating}
\usepackage{threeparttable}
\usepackage{booktabs}
\usepackage{ragged2e}
\usepackage{array}
\usepackage{url}
\usepackage{soul}

\begin{document}

% Copyright notice for posting on arXiv and personal website
\onecolumn

\textcopyright~ 2021 IEEE. Personal use of this material is permitted. Permission from IEEE must be obtained for all other uses, in any
current or future media, including reprinting/republishing this material for advertising or promotional purposes, creating new
collective works, for resale or redistribution to servers or lists, or reuse of any copyrighted component of this work in other
works.
\newpage
\twocolumn

\title{Spatio-temporal graph neural networks for multi-site PV power forecasting}

\author{Jelena~Simeunović, %~\IEEEmembership{Member,~IEEE,}
        Baptiste~Schubnel, %~\IEEEmembership{Fellow,~OSA,}
        Pierre-Jean~Alet
        and~Rafael~E.~Carrillo%~\IEEEmembership{Life~Fellow,~IEEE}% <-this % stops a space
\thanks{The authors are with CSEM, Neuch\^{a}tel, Switzerland (e-mails: jelena.simeunovic@csem.ch, baptiste.schubnel@csem.ch, pierre-jean.alet@csem.ch, rafael.carrillo@csem.ch). 
J. Simeunović is also with the Signal Processing Laboratory (LTS4), EPFL, Lausanne, Switzerland.}\vspace{-1.5em}% <-this % stops a space
%\thanks{Manuscript received April 19, 2000; revised August 26, 2015.}
\thanks{This work was carried out on behalf of and with the support of the Swiss Federal Office of Energy (research contract SI/501803-01). The authors are solely responsible for the content and conclusions of the paper. This work was also supported in part by BKW AG.}}%

\markboth{IEEE Transactions on Sustainable Energy, Accepted}%
{Simeunović \MakeLowercase{\textit{et al.}}: Spatio-temporal graph neural networks for multi-site PV power forecasting}

\maketitle
\begin{abstract}
Accurate forecasting of solar power generation with fine temporal and spatial resolution is vital for the operation of the power grid. However, state-of-the-art approaches that combine machine learning with numerical weather predictions (NWP) have coarse resolution. In this paper, we take a graph signal processing perspective and model multi-site photovoltaic (PV) production time series as signals on a graph to capture their spatio-temporal dependencies and achieve higher spatial and temporal resolution forecasts. We present two novel graph neural network models for deterministic multi-site PV forecasting dubbed the graph-convolutional long short term memory (GCLSTM) and the graph-convolutional transformer (GCTrafo) models. These methods rely solely on production data and exploit the intuition that PV systems provide a dense network of virtual weather stations. The proposed methods were evaluated in two data sets for an entire year: 1) production data from 304 real PV systems, and 2) simulated production of 1000 PV systems, both distributed over Switzerland. The proposed models outperform state-of-the-art multi-site forecasting methods for prediction horizons of six hours ahead. Furthermore, the proposed models outperform state-of-the-art single-site methods with NWP as inputs on horizons up to four hours ahead.
\end{abstract}

\begin{IEEEkeywords}
Photovoltaic systems, forecasting, machine learning, graph signal processing, graph neural networks 
\end{IEEEkeywords}

\IEEEpeerreviewmaketitle
\section{Introduction}

\IEEEPARstart{I}{mproving} power predictions of intermittent and non-dispatchable energy sources is one of the key elements that contribute to increasing the penetration of variable renewable energy in the power grid. Solar power depends on local weather conditions and cloud dynamics. The resulting high variability hampers the accuracy of its forecasts. State-of-the-art methods for PV forecasting combine machine learning methods with numerical weather predictions NWP but their resolution, both in time and space, might be too coarse, which poses a challenge for applications such as energy trading and grid congestion management \cite{alet_forecasting_2016}. 

Several approaches have been proposed to improve deterministic PV forecasting accuracy with higher spatial resolution. Previous works focusing on intra-day (up to six hours ahead) predictions use inputs from various sources, in particular: ground-based cameras \cite{article_chu},  satellite images \cite{7437475, scmidt}, and NWP \cite{ANTONANZAS201678}. Ground-based cameras are expensive to deploy and maintain in a grid with a large number of PV stations. Furthermore, they yield high accuracy only for intra-hour forecasts (up to one hour). On the other hand, satellite images are more suitable for regional forecasting when PV stations are clustered, since the wide-area images are inadequate for providing site-specific information. Methods that combine satellite images with NWP \cite{Huang, Li_2016, SPERATI2016437, marco_new} excel in  long-term forecasts, but they perform rather poorly at short-term horizons and high spatial resolution. Precise local numerical weather forecasts may be accessible by dedicated meteorological providers, but are often very costly to acquire and require heavy processing. Therefore, in order to avoid the issues brought by the additional exogenous data, the question arises whether it is possible to achieve state-of-the-art results relying only on past PV power generation data.

Different classes of machine learning models have previously been reported to investigate this question. Traditional approaches are based on auto-regressive (AR) linear models \cite{Yang2015MultitimeScaleDS}. These were further extended to vector auto-regressive (VAR), Lasso-VAR \cite{7981201, article_agoua}, graph-based spatio-temporal AR \cite{carrillo2020high} and auto-regressive moving average (ARM) models \cite{unknown}.  Simple nonlinear neural networks, however, outperform persistence model and simple linear methods for forecasting horizons longer than one hour \cite{lauret}. Nonlinear neural network models include recurrent and convolutional  neural networks.  The recurrent structures with long short-term memory (LSTM) network \cite{GhaderiSG17}, \cite{Lai_2018, Lee2018ForecastingSP} perform well at capturing temporal patterns. Since passing clouds influence neighboring PV sites sequentially, the cloud cover and the cloud movements can be captured by considering spatial and temporal relations between PV stations. For that purpose, convolutional neural networks (CNN) have been proposed to extract the spatio-temporal correlations by stacking the PV signals as an image and reordering their position in the image based on their location \cite{Jeong_2019, Zhu_2018}. In addition, attention mechanisms have been also introduced to capture spatial correlations \cite{zhou, shih}.  One of the main advantages of \cite{shih} is the high accuracy for different spatio-temporal forecasting tasks including electricity, PV, exchange rate and traffic forecasting, without tailoring the model to a specific task. Spatio-temporal forecasting models have been mainly applied for the traffic speed forecasting problem. LSTMs have been used to capture temporal correlations, while convolutional and attention structures have been proposed to capture spatial relations \cite{Lai_2018, shih}. Although these models are complex, they use only a limited number of data steps from previous days as input to the model, thus, neglecting the temporal shift and periodicity. Therefore, bidirectional LSTMs have been proposed to exploit not only forward dependencies but also backward dependencies \cite{cui2018deep}. Bidirectional LSTM structures have also been used in \cite{9310216} to improve the probabilistic forecast of distribution locational marginal prices by accessing long-term dependencies. 
One drawback of the latter work is that the spatial information needs to be carefully encoded and concatenated as features to the input data and the bidirectional LSTM is used to implicitly learn the spatial relations.
 
Although the aforementioned works exploit spatio-temporal correlations, they do not fully exploit the spatial information of multiple  sites. Graph signal processing (GSP) is an emerging field that allows the processing of signals defined in irregular domains by using graphs to capture their interdependence  \cite{8347162}. Recently, graph neural networks (GNN) have attracted a lot of attention due to their expressive power and ability to infer information from complex data such as brain signals, social networks interactions and traffic congestion patterns \cite{zhou2019graph, zonghan}.  Spatio-temporal graph forecasting has been studied in various fields, including traffic forecasting, weather forecasting and price forecasting among others. Spatio-temporal techniques in the traffic speed forecasting use gated graph convolutional structures \cite{Yu_2018} and encoder-decoder recurrent diffusion convolution \cite{li20172018diffusion} to capture spatial and temporal correlations. However, these models only predict one step ahead in one iteration and then use predictions as historical observations, which increases the error for longer-term predictions. This problem was addressed in \cite{8903252, zheng2019gman} that use attention mechanisms for multi-step prediction. However, traffic speed forecasting has a predefined graph topology, constructed from the road network, which makes the problem easier in comparison to PV or wind speed forecasting, where the correlations and connections between PV (or wind) systems are not known in advance. LSTMs coupled with graph convolutional structures for capturing spatio-temporal patterns were recently proposed for wind speed forecasting \cite{8650287}. This mechanism requires a different LSTM network for each site to learn the temporal relations followed by graph convolutional layers to learn the spatial dependencies. The main drawback of this approach is that it requires one model for each step ahead in the prediction horizon, which is not sample efficient (needing around four years of data for training) and not scalable for a large number of nodes. Graph models have also been used to produce probabilistic forecasts for solar irradiance in \cite{8663347}, which proposed a graph convolutional auto-encoder to model the irradiance's probability distribution at node level in a scalable fashion. 

In this paper, we take a GSP perspective and model the PV production time-series as signals on a graph. The intuition behind this choice is that for a sufficiently dense network of PV systems, graph-based models can exploit the spatio-temporal dependencies of PV production data to infer part of the cloud dynamics and forecast production more accurately. We present two novel spatio-temporal GNN models for  deterministic   multi-site PV power forecasting which rely entirely on production data: the Graph-Convolutional Long Short Term Memory (GCLSTM) and the Graph-Convolutional Transformer (GCTrafo) models. Both models use graph convolutional layers to infer the spatial patterns from the data though they use different structures to model the time dependence: GCLSTM uses recurrent structures, whereas GCTrafo uses attention mechanisms. The proposed models are compared with state-of-the-art methods for  deterministic  multi-site PV forecasting, for a forecasting horizon of six hours ahead, over an entire year in two datasets distributed over Switzerland: (1) production data from 304 real PV systems, and (2) simulated production of 1000 PV systems. Additionally, the proposed forecasting models are compared with single-site state-of-the-art forecasting methods that use NWP as inputs for two sites also in Switzerland.

The rest of the paper is organized as follows. Section \ref{section_2} introduces preliminaries on graph convolution and graph time series forecasting of PV generation. Section \ref{section_3} details the proposed GCLSTM and GCTrafo GNN architectures. Experimental results of our evaluation are presented and discussed in Section \ref{section_4}. Finally, we conclude in Section \ref{section_5}.

\section{Problem formulation}
\label{section_2}
\subsection{Graph convolution}
We begin by briefly reviewing some relevant concepts in graph signal processing. A weighted undirected graph $G$ is represented as a tuple $G=( \nu,\upvarepsilon, \bold{A})$, where $\nu = \{v_1,v_2,…,v_N \}$  is its set of vertices (nodes) and $\upvarepsilon$ its set of edges (links). If nodes $v_i$ and $v_j$ are connected, the edge between $v_i$ and $v_j$  is denoted by $e_{ij} \in\upvarepsilon $. The topology of the graph is determined by its symmetric adjacency matrix $\bold{A}$ of size $N \times N$. The matrix element $A_{ij}$ gives the edge weight between vertices $v_i $ and $v_j$ and is zero in the absence of an edge $e_{ij}$.  In the multi-site PV case, each PV station corresponds to a node in the graph $G$ and edges might represent the spatial proximity between the PV stations $v_i $ and $v_j$ or other relationship between stations.  The Laplacian matrix $\mathscr{\bold{L}}$ is defined by $\mathscr{\bold{L}}:= \bold{D}-\bold{A}$, where $\bold{D}$ is the diagonal matrix of nodes' degrees $ D_{ii}=\sum_j A_{ij}$. Being positive semidefinite, $\bold L = \bold{U} \boldsymbol{\Lambda}  \bold{U}^T$, where $\bold U$  is a unitary matrix of eigenvectors  and $ \boldsymbol{\Lambda} \in \mathbb R^{N \times N} $ is the diagonal matrix of associated eigenvalues $\lambda_i$, $i=1, \ldots,N$. Finally, we define a graph signal as a mapping $\mathscr{\bold{x} : \nu \rightarrow \mathbb{R}}$, such that ${x_v} \in \mathbb{R}$ is the signal value at node $v$.  In our case the graph signal $\bold{x}$ represents the vector containing the power production of all PV stations at some point time.  The graph Fourier transform of a signal $\bold{x}$ is defined as $\bold{\dot{x}}=\bold{U^T} \bold{x}$ and its inverse as $\bold{x}=\bold{U}\bold{\dot{x}}$.  For more details and in-depth review of GSP we refer the reader to \cite{8347162}. 

Graph convolutions  can be divided into two categories: spectral convolution \cite{8347162, bruna2014spectral, Defferrard2016ConvolutionalNN} and spatial graph convolution \cite{duvenaud2015convolutional, NIPS2016_390e9825}. The spectral graph convolution is defined in the graph Fourier domain: for a real function $h$, the graph convolution of a signal $\bold{x}$ with  $h$ is defined by
 \begin{equation}\label{eq:cf}
   h*_\mathscr{G} \bold{x} :=\bold{U} \bold{h}(\boldsymbol{\Lambda}) \bold{U^T} \bold{x},
 \end{equation}
where $\bold h(\boldsymbol{\Lambda})$ is the diagonal matrix with entries $h(\lambda_i) \in \mathbb{R}$, $i=1,...,N$.  In \cite{Defferrard2016ConvolutionalNN}, the authors use Chebyshev series expansion together with the scaling of the Laplacian eigenvalues to parametrize and approximate $\bold h(\boldsymbol{\Lambda})$ as:
\begin{equation}\label{eq:Cheb}
\bold{h(\boldsymbol{\Lambda})} \approx \sum_{k=0}^{K-1} \theta_k \bold{T}_k(\tilde{\boldsymbol{\Lambda}}),
\end{equation}
where $ \theta_k \in \mathbb{R}$ are  Chebyshev coefficients, $\tilde{\boldsymbol{\Lambda}}=  2 \boldsymbol{\Lambda}/ \lambda_{max} - \boldsymbol{I_N}$ is the scaled eigenvalue matrix, and $\bold{T}_k(\tilde{\boldsymbol{\Lambda}}) \in \mathbb R^{N \times N}$ is the diagonal matrix with diagonal entries the Chebyshev polynomial of order $k$ applied to the scaled eigenvalues. Using functional calculus and plugging into \eqref{eq:cf}, one finally gets
\begin{equation}\label{eq:conv}
h *_\mathscr{G}  \bold{x} \approx \sum_{k=0}^{K-1} \theta_k \boldsymbol{T_k(\tilde{L}})\bold{x},
\end{equation}
where  $\tilde{\boldsymbol{L}}=  2 \boldsymbol{L}/ \lambda_{max} - \boldsymbol{I_n}$ is the scaled Laplacian.  The main practical advantage of the right side in   \eqref{eq:conv} is to reduce the computation complexity from $\mathcal{O}(N^2)$ to $\mathcal{O}(K \vert \upvarepsilon \vert)$. Moreover, the graph convolutional filter represented  with polynomials of order K of the scaled Laplacian is spatially localized and only depends on nodes that are $K$-hops away from the central node. See \cite{Defferrard2016ConvolutionalNN} for a detailed discussion. In the rest of this paper, we  consider filters $h$ for which an equality in \eqref{eq:conv} holds, and make the abuse of notation
$$\theta *_\mathscr{G}  \bold{x} \equiv h *_\mathscr{G}  \bold{x}.$$ 

Moreover, we consider the extension of \eqref{eq:conv}  to multivariate graph signals ${\bold{x}_v} = (x_{v}^{(1)}, \ldots, x_{v}^{( \rm{f_{in}})})  \in \mathbb{R}^{\rm{f_{in}}}$  and replace $\theta \in \mathbb{R}^{K}$ by  $\bold W \in \mathbb R^{K  \times \rm{ f_{out}} \times \rm{f_{in}}}$, where $f_{in}$ and $f_{out}$ denote the number of input and output features, respectively, in the layer. Thus, for any  output feature  $j$ in $\rm{ f_{out}}$,
\begin{equation}\label{eq:conv:multivar}
(\bold W *_\mathscr{G}  \bold{x})^{(j)} = \sum_{k=0}^{K-1} \sum_{i=0}^{ \rm{f_{in}}} W_{k}^{(ji)} \boldsymbol{T_k(\tilde{L}})\bold{x}^{(i)}.
\end{equation}
The neural network architectures presented in this paper use in some layers the operation defined in \eqref{eq:conv:multivar}, the weights $\bold W$ being learnable parameters as in standard CNN.

\subsection{Multi-site time-series forecasting on graphs}
Atmospheric clouds act as a dynamic mask that affects local PV power production. For a sufficiently dense network of PV plants,  parts of this dynamics (diffusion and advection) can be inferred from past production data and used to predict future production on the entire network. Suppose we have $N$ PV stations. Thus, each station corresponds to a node in the network graph and the observed PV data are temporal signals attached to each node. The edge weight between two nodes is a measure of the expected correlation between two sites. Typical choices are bivariate (Pearson) correlation, distance correlation and different kernel-based methods \cite{Kriege_2020}. 

Let $\bold{p}(t) \in \mathbb R^N$ denote the vector of PV power production over all PV stations at time step $t$ with the value at  node $v$ being denoted by $p_v(t)$. Formally, we want to forecast $\bold{p}(t)$ for the next $H$ discrete time steps ahead given $M$ past observations as:
\begin{equation}\label{eq:Power_gsp}
\bold{\hat p}(t), \ldots, \bold{\hat{p}}(t+H-1)=f_\beta \left( \bold{p}(t-M),\dots, \bold{p}(t-1)\right),
\end{equation}
where $f_\beta$ is a chosen family of parametric estimators. The learning problem consists in finding a set of parameters $\beta$ that minimizes the prediction error over the entire horizon by solving 
 \begin{equation}\label{eq:parametric_model}
 \arg\min_\beta  \sum_{t \in \mathcal{T}}\sum_{\tau=t}^{t+H-1} \| \bold{\hat p}(\tau) - \bold{ p}(\tau) \|^{2}_{2},
\end{equation}
where $\mathcal{T}$ is the set of times of past observations taken into consideration to fit the model (training set).

\section{Graph convolutional forecasting models}
\label{section_3}
In this section we present two  sequence-to-sequence forecasting models based on spectral graph convolutions. Both share the same structure: an  encoder  to  process the past $M$ observed data and a decoder to predict the next $H$ future  observations. 

\subsection{Graph convolutional long-short term memory neural network}
\label{sec:gclstm}
\begin{figure}[t!]
  \begin{center}
   \includegraphics[trim=0.0cm 0.9cm 0.0cm 1.9cm, clip,width=\linewidth]{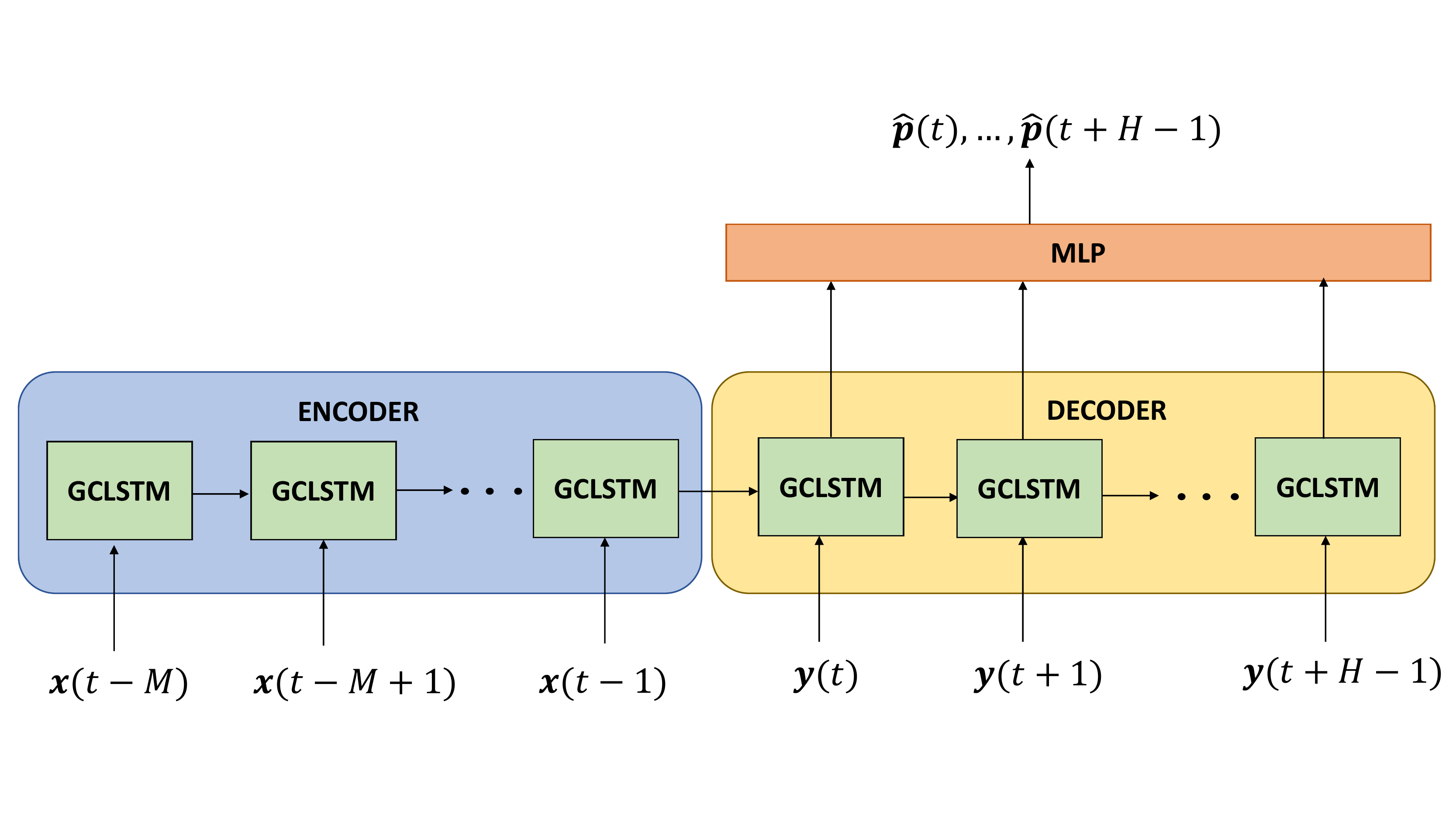}
    \caption{ Encoder-decoder Graph convolutional LSTM architecture.}
    \label{fig:gclstm}
\end{center}
\end{figure}

The first architecture is a sequence to sequence model based on graph convolutional long-short term memory (GCLSTM) (see \cite{lstm_art} for LSTM networks, \cite{bresson} for graph convolutional recurrent networks).  Both the encoder and decoder combine recurrence and spectral graph convolution to model jointly  temporal and spatial correlations. The encoder is a GCLSTM network that estimates the state of the system, given a sequence of  past observations, its initial state being set to zero. The decoder is another GCLSTM cell that is initialized with the final encoder state and predicts the power for the chosen horizon  period of $H$ steps ahead; see Figure \ref{fig:gclstm}. A multi-layer perceptron (MLP) is used at the output of the decoder to transform the GCLSTM outputs into the desired power production $\bold{\hat p}(\tau)$, where $\tau \in \{ t , \ldots, t+H-1\}$. The specific inputs features $\bold x (\tau)$ and $\bold y(\tau)$,  concatenations of  power and clear sky irradiance signals,  are presented at the end of the section.

The usage of LSTM cells as  recurrent structures of the model is justified by their capacity of learning  and retaining both short - and long-term dependencies; see \cite{lstm_art}. We denote by $\rm{lat}$ the number of dimensions of the LSTM cell latent representation. In the classical LSTM cell, the cell state $\bold{c}(\tau) \in \mathbb{R}^{\rm{lat}}$ and the output $\bold{h}(\tau) \in \mathbb{R}^{\rm{lat}}$  are updated recursively from the input sequence $\bold x(\tau) \in \mathbb{R}^{\rm{f_{in}}}$ using gating operations involving matrix multiplications. In GCLSTM cells, $\bold{c}(\tau) \in \mathbb{R}^{ N \times \rm{lat}}$,  $\bold{h}(\tau) \in \mathbb{R}^{ N \times \rm{lat}}$, $\bold x(\tau) \in \mathbb{R}^{N \times \rm{f_{in}}}$ and the gating operations are modified by replacing the matrix multiplications with spectral graph convolutions as defined in \eqref{eq:conv:multivar}. Doing so,  signals  are diffused across neighboring nodes and local spatial information is better captured (see \cite{Defferrard2016ConvolutionalNN} ,\cite{seo2018structured}). For a given input sequence $(\bold x(\tau))_{\tau}$, the GCLSTM cell equations are given by
\begin{equation}\label{eq:lstm}
\begin{split}
\bold f(\tau) &= \sigma(\bold W_{f,h} *_\mathscr{G} \bold h(\tau-1) + \bold W_{f, \rm{x}} *_\mathscr{G} \bold x(\tau) +\bold b_f)\\
\bold i(\tau) &= \sigma(\bold W_{i,h} *_\mathscr{G}   \bold h(\tau-1) + \bold W_{i,\rm{x}} *_\mathscr{G} \bold x(\tau) +\bold b_i)\\
\bold o(\tau) &=\sigma(\bold W_{o,h} *_\mathscr{G} \bold h(\tau-1) + \bold W_{o,\rm{x}} *_\mathscr{G} \bold x(\tau) +\bold b_o)\\
\bold c(\tau) &= \bold  i(\tau) \otimes  \tanh(\bold W_{c,h} *_\mathscr{G} \bold h(\tau-1)+ \bold W_{c,\rm{x}} *_\mathscr{G} \bold x(\tau) + \bold b_c)\\
&+ \bold f(\tau) \otimes \bold c(\tau-1)\\
\bold h(\tau) &= \bold o(\tau) \otimes \tanh(\bold c(\tau))
\end{split}
\end{equation}

where $\sigma$ is the sigmoid function, $ \bold{W} *_\mathscr{G} \cdot$ is defined in \eqref{eq:conv:multivar} and $\otimes$ is the Hadamard product. The dimension of the weights $\bold{W}_{\cdot, \cdot}$ and biases $\bold{b}_{\cdot}$ are determined by the number of dimensions of the input feature space, $\rm{f_{in}}$,  the latent space,  $\rm{lat}$, and the order $K$ of the Chebyshev expansion: $W_{\cdot,h} \in \mathbb{R}^{K \times \rm{lat} \times \rm{lat}}$ and $W_{\cdot,\rm{x}} \in \mathbb{R}^{K \times \rm{lat} \times \rm{f_{in}}}$, the biases being in $\mathbb{R}^{K \times \rm{lat}}$. 

The adjacency  matrix $\bold{A}$ of the graph is initialized using the $k$-nearest neighbors algorithm:  $A_{ij} =1$ if $v_{i}$ and $v_j$ are nearest neighbors, and 0 otherwise. The scaled Laplacian $\bold{\tilde{L}}$ involved in the graph convolutions in \eqref{eq:lstm} is calculated initially from  $\bold{A}$ and represented as a sparse tensor. In the course of training, not only weights and biases in \eqref{eq:lstm} are learnt, but also the non-zero entries of the sparse Laplacian in each cell operation, so as to capture very local specifics related, for instance, to the topology of the terrain or nodes separation distance. Encoder and decoder are trained simultaneously.

The input sequence of the encoder consists of tuples $\bold{x}(\tau) = (\bold{p}(\tau), \bold{ \bar p}(\tau),\bold{g}(\tau))$, $\tau \in \{t-M, \ldots, t-1\}$, where $\bold{p}(\tau) \in \mathbb R^N$ is the power produced at time $\tau$, $\bold{g}(\tau) \in \mathbb R^N$  the global clear sky irradiance at time $\tau$ and $\bold{\bar p}(\tau) \in \mathbb{R}^N$ is the rolling mean power produced over the interval $[ \tau -72h, \tau-24h ]$. The clear sky irradiance values are computed at any location and any time on the map using the Ineichen and Perez clear sky model from PVlib \cite{stein2016pvlib}. This computation is deterministic and only relies on the geographical coordinates of the nodes (latitude, longitude and altitude). Similarly, inputs to the decoder are sequences of $\bold{y}(\tau) = (\bold{g}(\tau), \bold{d}(\tau),\bold{ \bar p}(\tau))$, where $\bold{d}(\tau)$ is the direct clear sky irradiance at time $\tau \in \{  t , \dots, t+H-1\}$. 

\subsection{Graph convolutional transformer}
\label{sec:gctrafo}
Even if the gate operations in \eqref{eq:lstm} protect the cell state $\bold{c}(\tau)$ and allow it to keep information over time, this information has the tendency to fade and be diluted  \cite{Schoene2020BidirectionalDL}. This shortcoming has been addressed in recent natural language processing architectures, in particular in the sequence to sequence model presented in \cite{10.5555/3295222.3295349}, called transformer. In transformers,  access to past signals at any time is guaranteed thanks to the usage of dot-product attention between any two elements of the time sequence. The second architecture presented in this paper, dubbed graph convolutional transformer (GCTrafo), is inspired by the base transformer architecture but incorporates a slight number of modifications so as to make it suitable for the multi-site PV generation forecasting problem. As the GCLSTM presented in Section \ref{sec:gclstm}, the GCTrafo architecture is made of an encoder to process past signal values and a decoder to predict the future outcomes. Moreover, it shares the same encoder input, output and decoder output signals. However, the inner operations are quite different and the GCTrafo does not incorporate any recurrent structure. 

\begin{figure*}[h!]
  \begin{center}
   \includegraphics[width=0.9\linewidth]{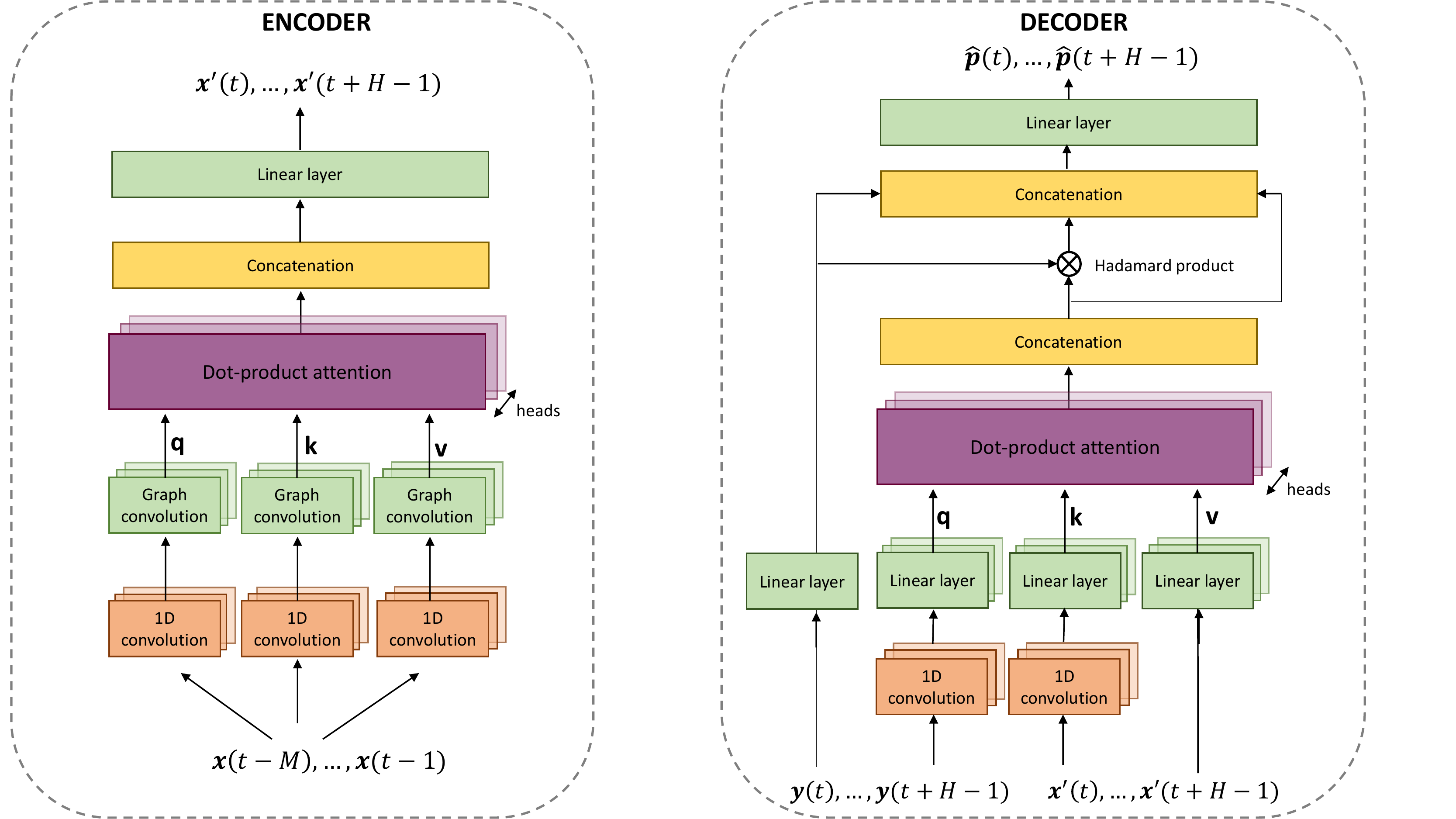}
    \caption{Graph Convolutional Transformer architecture.}
    \label{fig:gctrafo}
\end{center}
\end{figure*}

The encoder consists of three main stages, see Figure \ref{fig:gctrafo}. At the first stage, a 1D-convolutional layer is applied to the input sequence $(\bold x(\tau))_{\tau}$,  where $\tau \in \{t-M, \ldots, t-1\}$,  along the time axis to extract valuable variation features of the raw signals at single node level. This operation is made 3 times in parallel and produces three output sequences $(\bold{\tilde{x}}(\tau))_{\tau}$, $(\bold{\breve{x}}(\tau))_{\tau}$ and $(\bold{\check{x}}(\tau))_{\tau}$. At the second stage, a dot  product attention mechanism preceded by graph convolutions is used to embed every node variation feature in its spatio-temporal context. Queries $\bold q(\tau) \in \mathbb{R}^{N \times \rm{lat}}$, keys $\bold k(\tau)\in \mathbb{R}^{N \times \rm{lat}}$ and values $\bold v(\tau) \in \mathbb{R}^{N \times \rm{lat}}$ signals are first created by graph convolutional layers based on \eqref{eq:conv:multivar}, extracting neighboring node information: 
\begin{equation}
\begin{split}
\label{eq:conv:trafo}
\bold{q}(\tau) &= \bold{W_q} *_\mathscr{G} \bold{\tilde{x}}(\tau), \\
\bold{k}(\tau) &= \bold{W_k} *_\mathscr{G} \bold{\breve{x}}(\tau), \\
\bold{v}(\tau) &= \bold{W_v} *_\mathscr{G} \bold{\check{x}}(\tau).
\end{split}
\end{equation}

Queries, keys and values are then fed to a softmax dot product attention layer (see \cite{10.5555/3295222.3295349}), given for the query $\bold{q}(\tau)$ at  time $\tau$ and node $v$ by
\begin{equation}
\label{eq:dot:prod}
\begin{split}
\bold{att} (\bold{q}(\tau), & (\bold{k}(\tau'))_{\tau'},(\bold{v}(\tau'))_{\tau'}  )_{v} =\\
&\sum_{\tau'}  \frac{\exp(\bold{q}_{v}(\tau) \cdot  \bold{k}_{v}(\tau') )}{\sum_{\tau''} \exp(\bold{q}_v(\tau) \cdot  \bold{k}_v(\tau'')) }  \bold{v}_{v} (\tau') \in \mathbb{R}^{\rm{lat}}.
\end{split}
\end{equation}

In \eqref{eq:dot:prod}, the dot product inside the exponents contracts the latent dimensions to produce a scalar for each node $v$. A multi-head approach similar to \cite{10.5555/3295222.3295349} is used: the 1D convolutions, graph convolutions and dot product attention layers are duplicated $n$ times and concatenated before being fed to the last linear layer, producing the encoded sequence $(\bold{x'}(\tau))_{\tau}$. The intuition behind multi-head attention mechanism is that each head will learn to focus at different patterns of the input sequence.

The decoder operations are similar to the encoder with specifics related to the input sequence of the decoder.  The main difference is the absence of graph convolution  in the decoder before the attention layer. Indeed, the input sequence of the decoder consists of clear sky irradiance values and rolling mean values whose propagation across neighboring nodes is not expected to  add any further useful information.  Moreover, the value $\bold{v}(\tau)$ is directly set to be equal to the encoded vector $\bold{x'}(\tau)$, without any prior layer mapping; see Figure \ref{fig:gctrafo}. 
Finally, viewing the output of the attention layer  of the decoder as a vector encoding shading information coming from the cloud dynamics, this vector is concatenated with an embedding of the input $\bold{y}(\tau)$  at time $\tau \in \{ t, \dots, t+H-1\}$   and its Hadamard product with this embedding  before the last linear layer.  The last layer produces the output power production $\bold{\hat p}(\tau)$, where $\tau \in \{ t , \ldots, t+H-1\}$. 

During training, we adopt a similar strategy as for the GCLSTM encoder-decoder: all weights are learnt by stochastic gradient descent, and the non-zero entries of the scaled  Laplacian operator entering in the convolutions in \eqref{eq:conv:multivar} are learnt during training, starting with values equal to the ones derived using the $k$-nearest neighbor algorithm for the adjacency matrix. 

\section{Experimental Results}
\label{section_4}
In this section, GCLSTM and GCTrafo architectures are evaluated on two datasets, for both multi-site and single-site forecasts. In the following we describe the experimental setting first and then present the results.

\begin{figure*}[h!]
  \centering
  \begin{subfigure}{0.5\linewidth}
    \includegraphics[width=\linewidth]{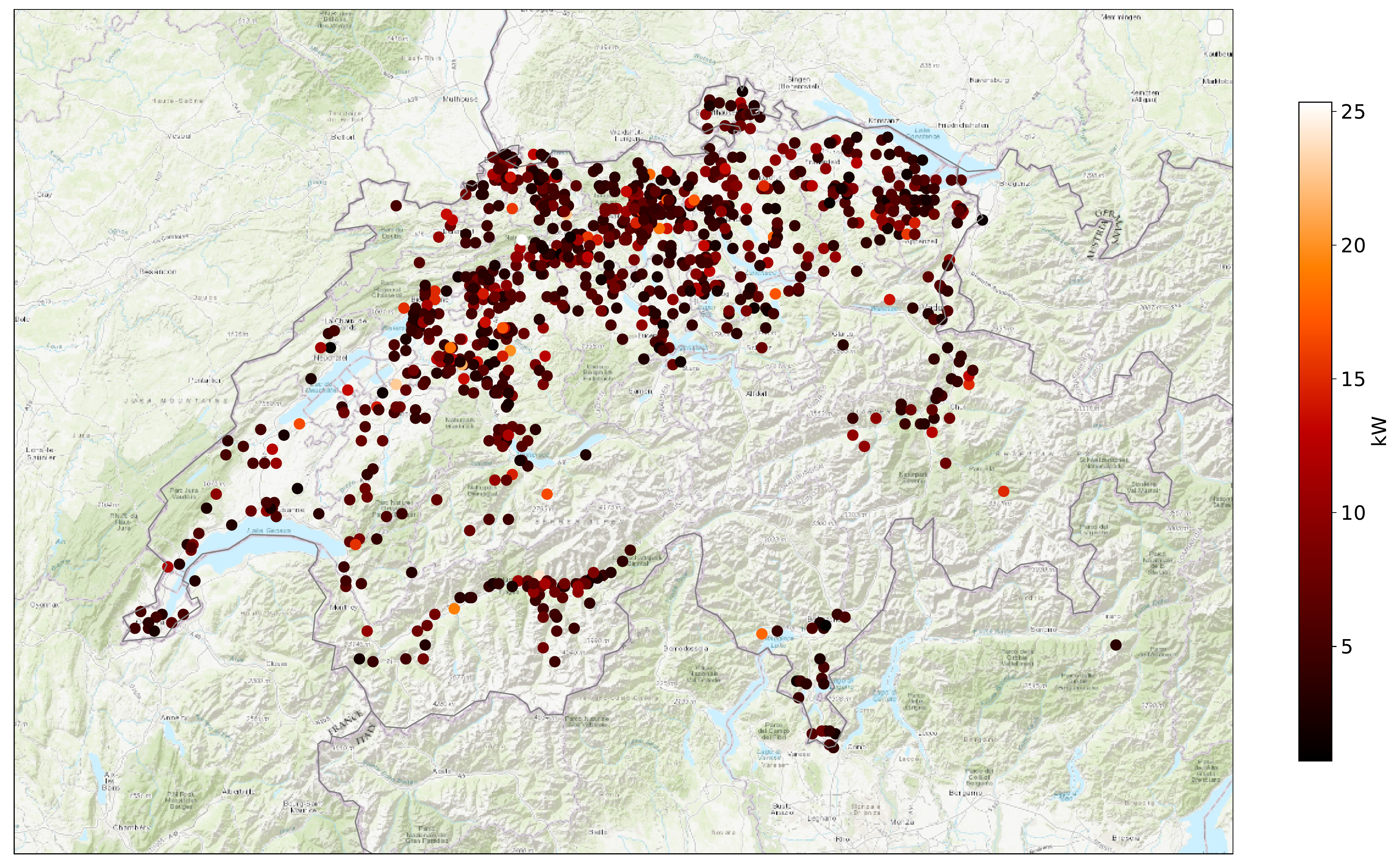}
    \caption{Synthetic dataset.}
      \label{fig:swiss_s}
  \end{subfigure}\hfill
  \begin{subfigure}{0.5\linewidth}
    \includegraphics[width=\linewidth]{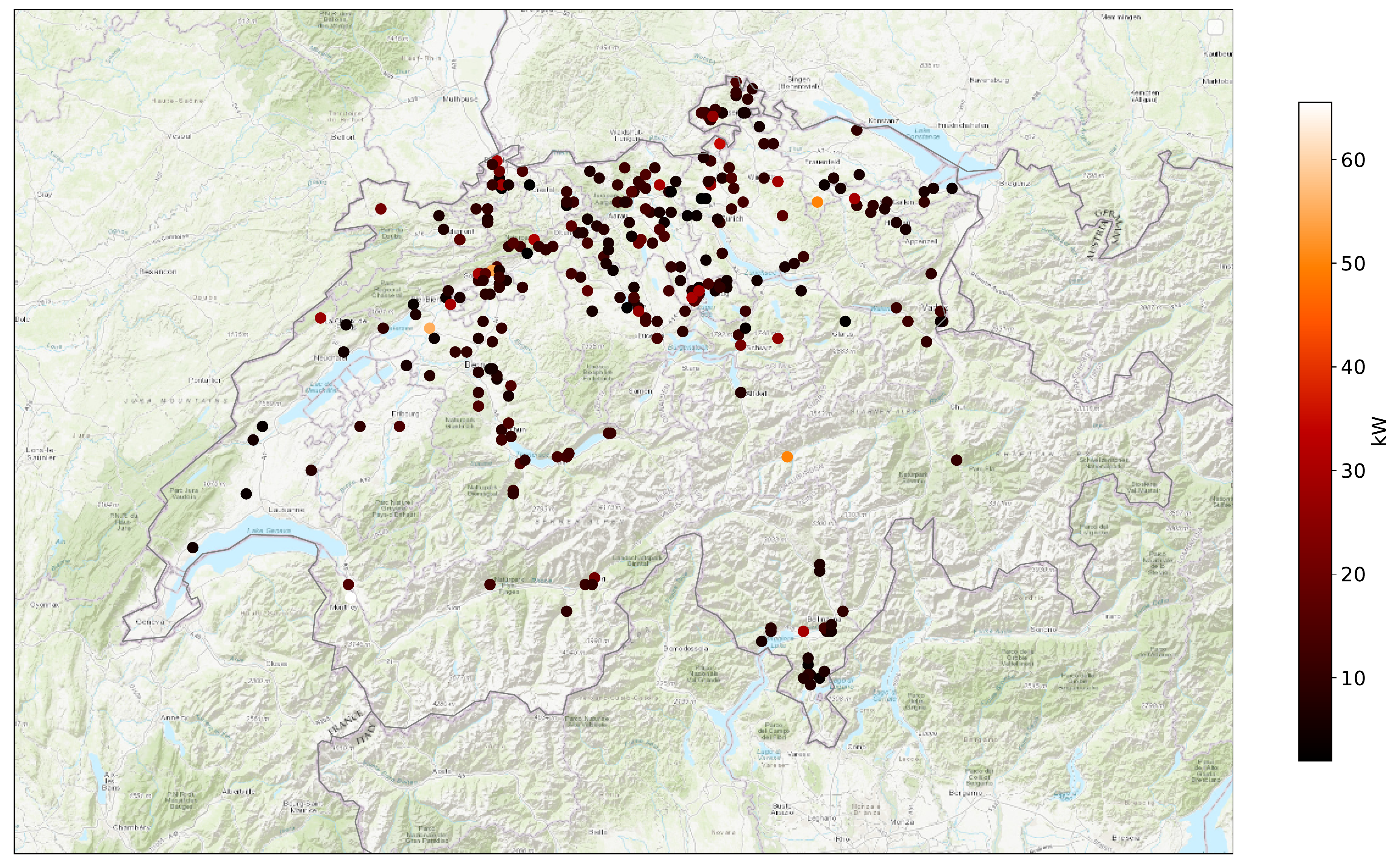}
    \caption{Real dataset.}
      \label{fig:swiss_r}
  \end{subfigure}
  \caption{ Spatial distributions of datasets. Colors indicate the peak production at each site.}
  \label{fig:swiss}
\end{figure*}

\subsection{Datasets}
Two datasets were used in our study. The first dataset, dubbed the real dataset, consists of records from 304 PV plants across Switzerland. The PV plants are spread inhomogeneously over the entire country, with a density reflecting the population density. The second dataset, dubbed the synthetic dataset, has 1000 PV plants and has been generated with statistical models that match the statistics of the real dataset in terms of location density, size, orientation and pitch angles. Production time series were simulated using the PVlib python library \cite{stein2016pvlib} and historical weather data from the HelioClim 3 database\footnote{\url{http://www.soda-pro.com/help/helioclim/helioclim-3-overview}}, with high temporal and spatial resolution, as inputs (see \cite{carrillo2020high} for further details). The spatial distribution of the real and synthetic dataset are shown in Figure \ref{fig:swiss}. Both datasets have a 15-minutes resolution for the years 2016-2018. 

The weather data used for the single-site benchmarks methods were obtained from two different meteorological providers. The forecast for Bern was computed using historical NWP from the global forecast system (GFS)\footnote{\url{https://www.ncdc.noaa.gov/data-access/model-data/model-datasets/global-forcast-system-gfs}} that has a temporal resolution of 3 hours. On the other hand, the forecast for Bätterkinden was computed using historical NWP from Meteotest\footnote{\url{https://meteotest.ch/en/}} with a temporal resolution of 1 hour.

\subsection{Baselines}
Two state-of-the-art methods were used as benchmarks in the multi-site forecasting evaluation. The first one is the recently proposed graph-based Spatio-temporal autoregressive model (STAR) \cite{carrillo2020high}. This method uses an AR model and a group Lasso estimator to select relevant plants (nodes) for the prediction of each individual site (node). The second baseline for multi-site forecasting is the non-graph-based Space-time convolutional neural network (STCNN) \cite{Jeong_2019}. It uses a greedy-adjoining algorithm that rearranges the plants based on their geographical proximity, one by one, before applying 2D convolution layers as in image processing to produce spatio-temporal features.

Apart from the benchmark methods used for the multi-site evaluation, in the single-site evaluation we used two state-of-the-art methods for single-site PV forecasting that use NWP. The first baseline for single-site comparison is a Support Vector Regression (SVR) model with NWP (global irradiance and temperature) as inputs. It was chosen as benchmark, since it was shown in \cite{boegli_ml_2018} that SVR outperforms several state-of-the-art methods for intra-day forecasts. The second single-site baseline is a state-of-the-art deep learning model, an Encoder-Decoder long-short term memory neural network (EDLSTM) \cite{mukhoty2019sequence}. It has a similar architecture to GCLSTM, such that both the encoder and decoder are based on LSTM networks. The decoder uses past observations of weather and PV site data to estimate the state of the system, and the decoder uses the state from the decoder as input as well as NWP (global irradiance and temperature) to forecast the site power. In addition to NWP data, EDLSTM uses the clear sky global irradiance for the site.  

\subsection{Data preprocessing}
Power data were normalized for both the real and synthetic dataset in the same manner for GCLSTM, GCTrafo and STCNN: The data for each node are normalized by the maximum power production over the training year. The STAR, albeit requires careful normalization in order to extract daily mode profiles from the past measurements. The tailor made normalization scheme is of utmost importance in the case of linear methods and for STAR is described in \cite{carrillo2020high}. 

The considered NWP for the SVR and EDLSTM models contains gaps and have a coarser resolution than the power data. In order to obtain 15-minutes resolution data without gaps, polynomial interpolation was used for  GFS data (Bern) and a sample-and-hold interpolation was applied to Meteotest data (Bätterkinden). All weather data were normalized before training using min-max scaling.

\begin{figure*}[h!]
  \centering
  \begin{subfigure}{0.49\linewidth}
    \includegraphics[trim=1.6cm 1.6cm 2.3cm 2.4cm, clip, width=\linewidth]{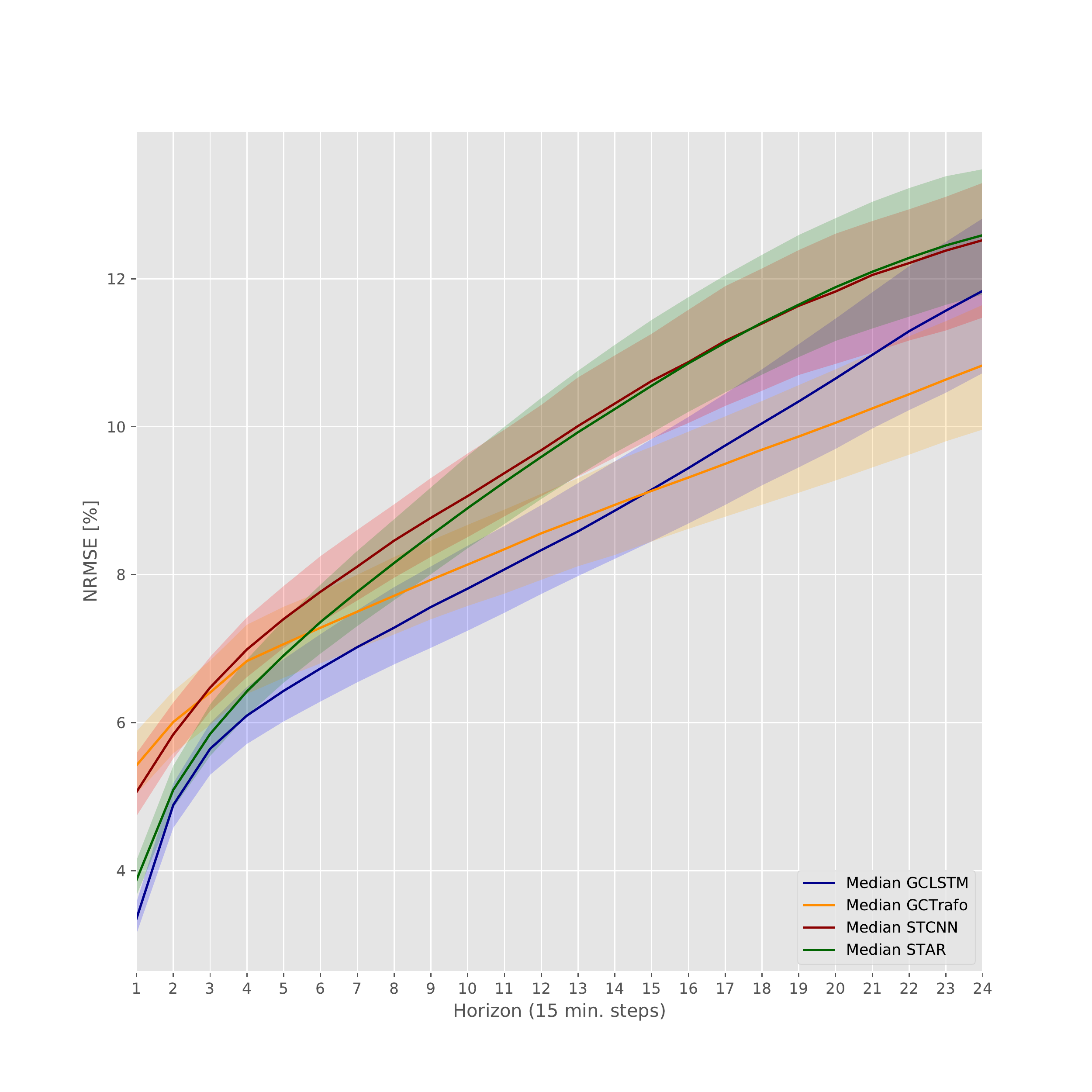}
    \caption{Forecast NRMSE for the synthetic dataset.}
     \label{fig:nrmse_s}
  \end{subfigure}\hfill
  \begin{subfigure}{0.49\linewidth}
    \includegraphics[trim=1.6cm 1.6cm 2.3cm 2.4cm, clip, width=\linewidth]{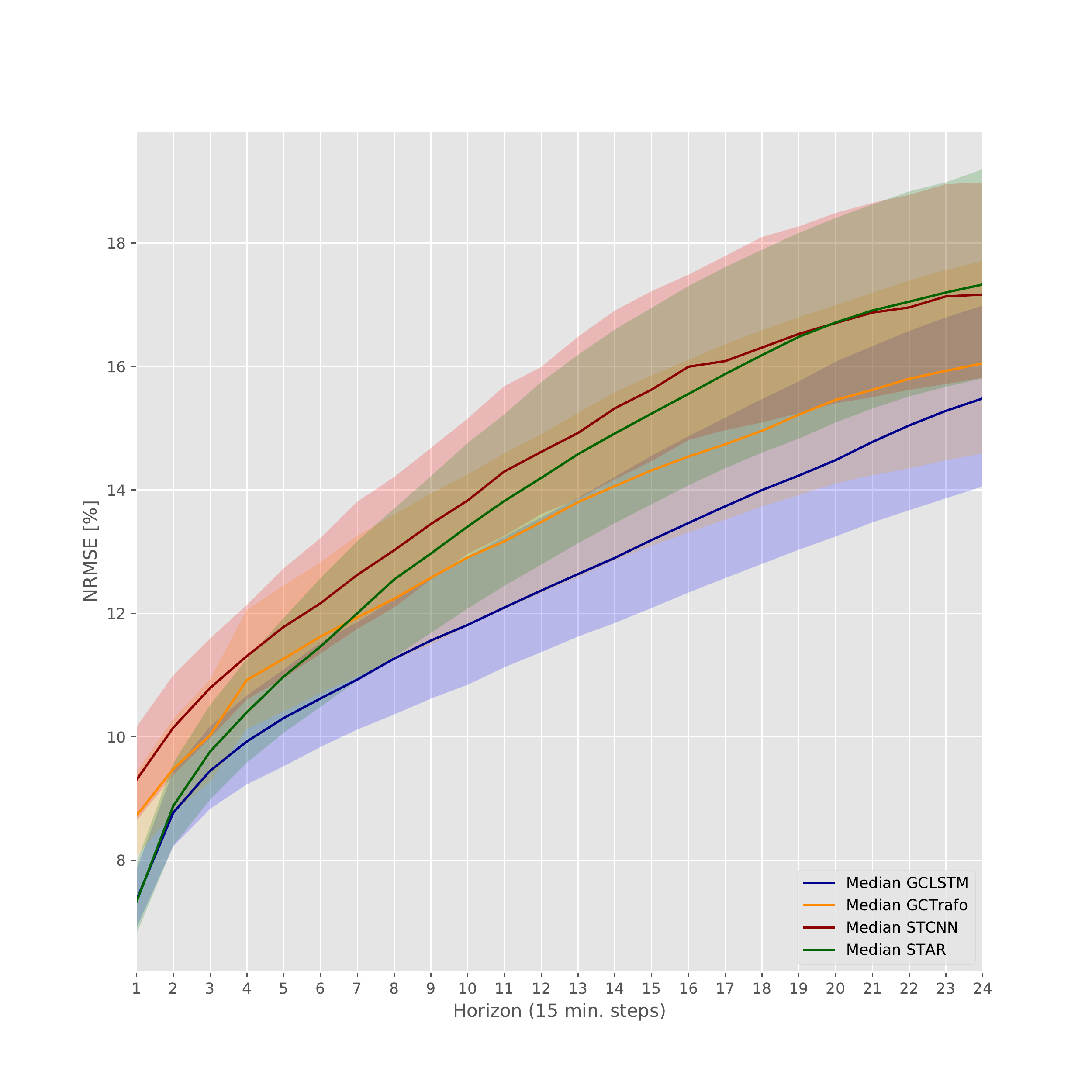}
     \caption{Forecast NRMSE for the real dataset.}
       \label{fig:nrmse_r}
  \end{subfigure}
  \caption{Forecast error comparison for multi-site PV power prediction. The forecast horizon is six hours in steps of 15 minutes. Solid lines show the median error while the shaded areas show the inter-quantile distance of the errors.}
\label{fig:nrmse}
\end{figure*}

\subsection{Training}
All methods, except STAR,  were trained on the first year of available data (2016) and evaluated on the second year (2017). STAR model coefficients were fitted over two months and then used to predict the power production over the next two weeks. The models were re-fitted every two weeks in a rolling window fashion for the entire 2017 year. The hyperparameters used in the developed and baseline models are described in more detail in the Appendix. All models were trained on a workstation with 16 cores, 128 GB of RAM memory and a Nvidia RTX 2080 Ti GPU.

\subsection{Evaluation and metrics}
The proposed models were compared over the year 2017 on the two datasets. Both the peak normalized root mean-squared error (NRMSE) and the averaged normalized mean absolute error (NMAE) are used as metrics. They are defined at site $v$ and forecasting step (horizon) $i$ as:
\begin{equation*}\label{eq:nrmse}
NRMSE(v, i) = \sqrt{\frac{1}{T} \sum_{t \in S} \left(\frac{\hat p_{v}(t+i)-p_{v}(t+i)}{{p^{max}_v}}\right)^2},
\end{equation*}
\begin{equation*}\label{eq:nmae}
NMAE(v, i)=  \frac{\sum_{t \in S} |\hat p_{v}(t+i)- p_{v}(t+i)|  }{\sum_{t \in S} p_{v}(t+i) }, 
\end{equation*}
where $p_v(t)$ and  $\hat{p}_v(t)$ denote the ground truth power and predicted power, respectively, of site $v$ at time $t$, $p^{max}_v$ is the maximum power of site $v$ over the evaluation period $S$, i.e., the 2017 year, and $T$ is the number of time steps in the evaluation interval $S$. Night times are excluded from error computations.

\subsection{Results}

We start by evaluating the performance of GCLSTM and GCTrafo on multi-site PV forecasting. Figures \ref{fig:nrmse_s} and \ref{fig:nrmse_r} show the evolution of the prediction errors of aforementioned methods over a horizon of 6 hours ahead, in steps of 15 minutes, for the synthetic and real dataset, respectively. The shaded regions represent the inter-quartile (25\%-75\%) error range over all nodes and solid lines represent the median NRMSE over all nodes. Results show that in the real dataset GCLSTM outperforms all other methods for the entire prediction horizon. However, in the synthetic dataset GCLSTM yields the lowest error up to 4 hours ahead, whereas GCTrafo outperforms the other models for predictions from 4 to 6 hours ahead. This shows the effectiveness of GNNs to capture the spatio-temporal correlations of the PV production data.  The synthetic dataset has a lower forecasting error due to the spatial and time smoothing in the generation of the HelioClim 3 irradiance database used to synthesize the PV power profiles. 

Although the linear STAR method performs better than the GCTrafo within the first hour on both datasets, GCTrafo shows
a lower error slope for horizons from one hour ahead on than the other methods, making it a promising model for longer prediction horizons, e.g. from 4 hours to day ahead. The main reason is that GCTrafo has the attention weights that can focus on different spatial or temporal information. Attention weights are more powerful than the recurrent structures, which suffer from fading memory for longer sequences.

\begin{table*}[t!]
\caption{Forecasting performance of proposed and baseline models on the two datasets}
\label{tab:error}
\resizebox{\textwidth}{!}{
\begin{tabular}{|c|c|c|c|c|c|c|c|c|c|c|c|c|c|c|c|c|}
\hline
        & \multicolumn{8}{c|}{Synthetic dataset}                                                                   & \multicolumn{8}{c|}{Real dataset}                                                                        \\ \hline
        & \multicolumn{2}{c|}{15min} & \multicolumn{2}{c|}{1h} & \multicolumn{2}{c|}{3h} & \multicolumn{2}{c|}{6h} & \multicolumn{2}{c|}{15min} & \multicolumn{2}{c|}{1h} & \multicolumn{2}{c|}{3h} & \multicolumn{2}{c|}{6h} \\ \hline
        & NRMSE        & NMAE        & NRMSE      & NMAE       & NRMSE      & NMAE       & NRMSE      & NMAE       & NRMSE       & NMAE         & NRMSE      & NMAE       & NRMSE      & NMAE       & NRMSE      & NMAE       \\ \hline
STAR    & 3.870      & 8.68     & 6.42     & 14.87   & 9.59     & 22.98   & 12.59    &  \textbf{28.52}                                                                              & \textbf{7.32}      & 15.80      & 10.40    & 23.83   & 14.20   & 33.15   & 17.33    & 40.41    \\ \hline
GCLSTM  & \textbf{3.350}     &  \textbf{7.23}      &  \textbf{6.09}    &  \textbf{13.32}    &  \textbf{8.33}     &\textbf{ 20.49}    & 11.84    &29.14               & 7.36      & \textbf{15.71}      &  \textbf{9.93}     &  \textbf{22.48}   &  \textbf{12.40}   & \textbf{29.36}    &  \textbf{15.53}    & \textbf{ 39.44}   \\ \hline
GCTrafo & 5.420      & 18.75     & 6.83    & 21.42    & 8.55     &  24.65    &  \textbf{10.83}    &  29.89                                & 8.76      & 20.27     & 10.95    & 25.93    & 13.54    & 33.13    & 16.07    & 40.50   \\ \hline
STCNN   & 5.060      & 13.63     & 6.99     & 17.89    & 9.68    & 25.06   & 12.52    & 31.97                                                              & 9.30      & 21.91      & 11.31  & 26.98    & 14.62    & 37.22    & 17.17    & 43.85 \\ \hline
\end{tabular}}
\end{table*}

\begin{figure}[h!]
  \begin{center}
     \includegraphics[trim=0.3cm 0.3cm 0.3cm 0.5cm, clip, width=\linewidth]{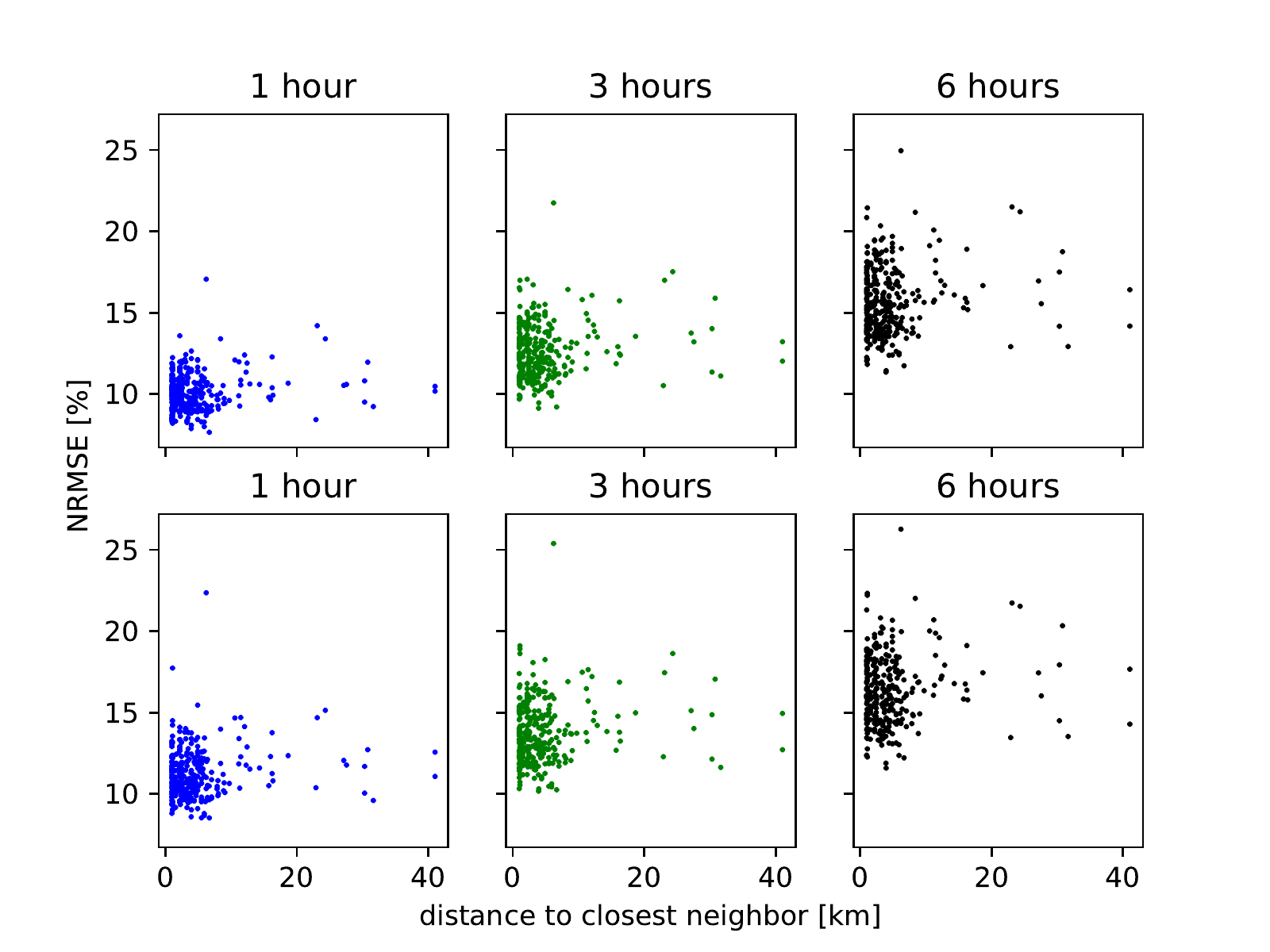}
    \caption{ NRMSE with respect to the distance to the closest neighbor for 1, 3 and 6 hours predictions for GCLSTM (top) and GCTrafo (bottom). }
    \label{fig:knn1}
\end{center}
\end{figure}

The comparative NRMSE and NMAE on both datasets are shown in Table \ref{tab:error} for 15 minutes, 1 hour, 3 hours and 6 hours ahead predictions. NRMSE is more sensitive to outliers, because it considers squared errors, therefore, gives more weight to large errors, thus using both metrics is useful when comparing methods. For example, this difference is highlighted in the errors for  6 hours ahead in the synthetic dataset. Although GCLSTM has a lower NMAE, the NRMSE is lower for GCTrafo.  Also, the NMAE gives a figure of the forecast error in percentage of the total yearly production.

\begin{figure*}[h!]
  \centering
  \begin{subfigure}{0.45\linewidth}
    \includegraphics[trim=1cm 0.5cm 1.9cm 2cm, clip, width=\linewidth]{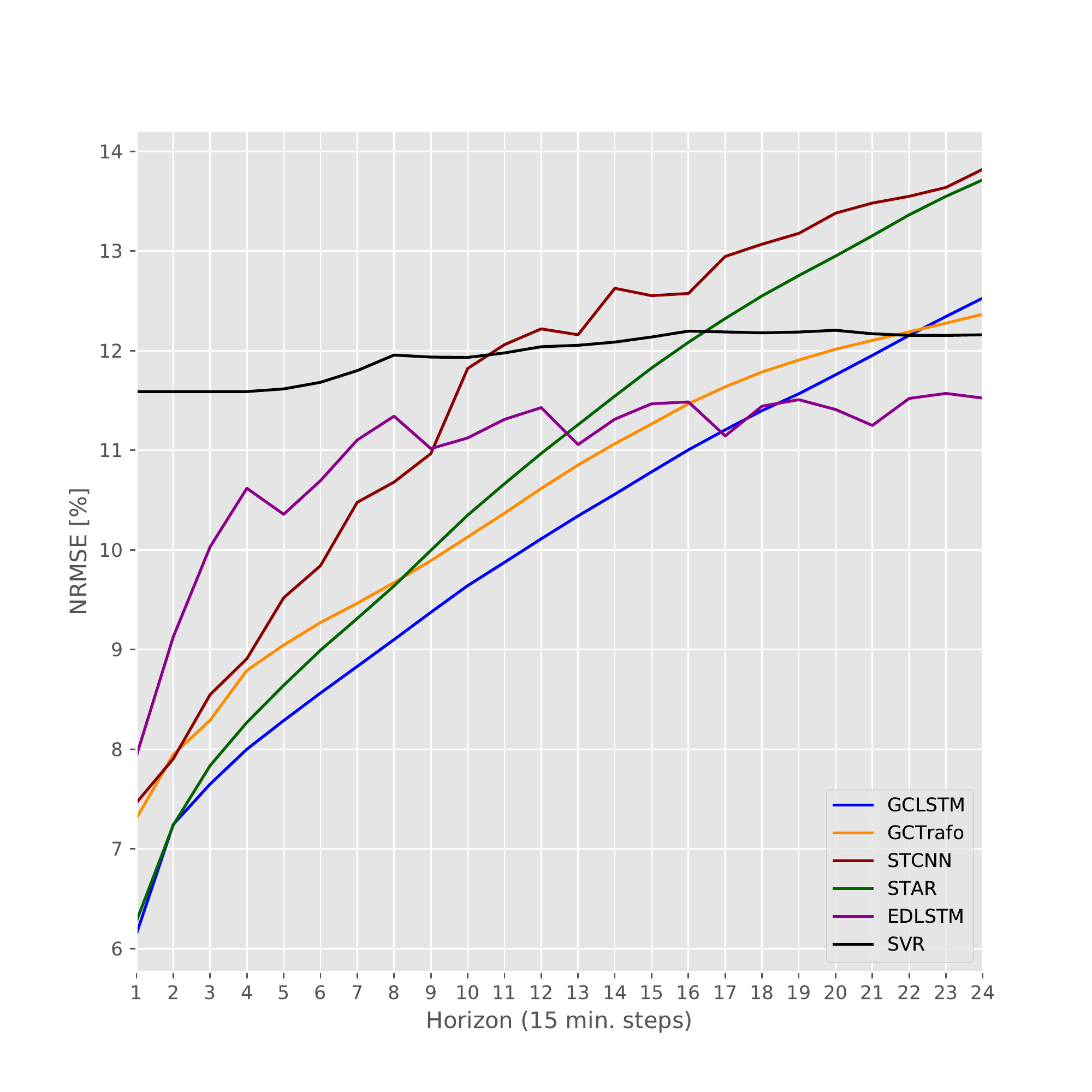}
    \caption{Forecast NRMSE for Bätterkinden.}
     \label{fig:nrmse_batterkinden}
  \end{subfigure}\hfill
  \begin{subfigure}{0.45\linewidth}
    \includegraphics[trim=1cm 0.5cm 1.9cm 1.8cm, clip, width=\linewidth]{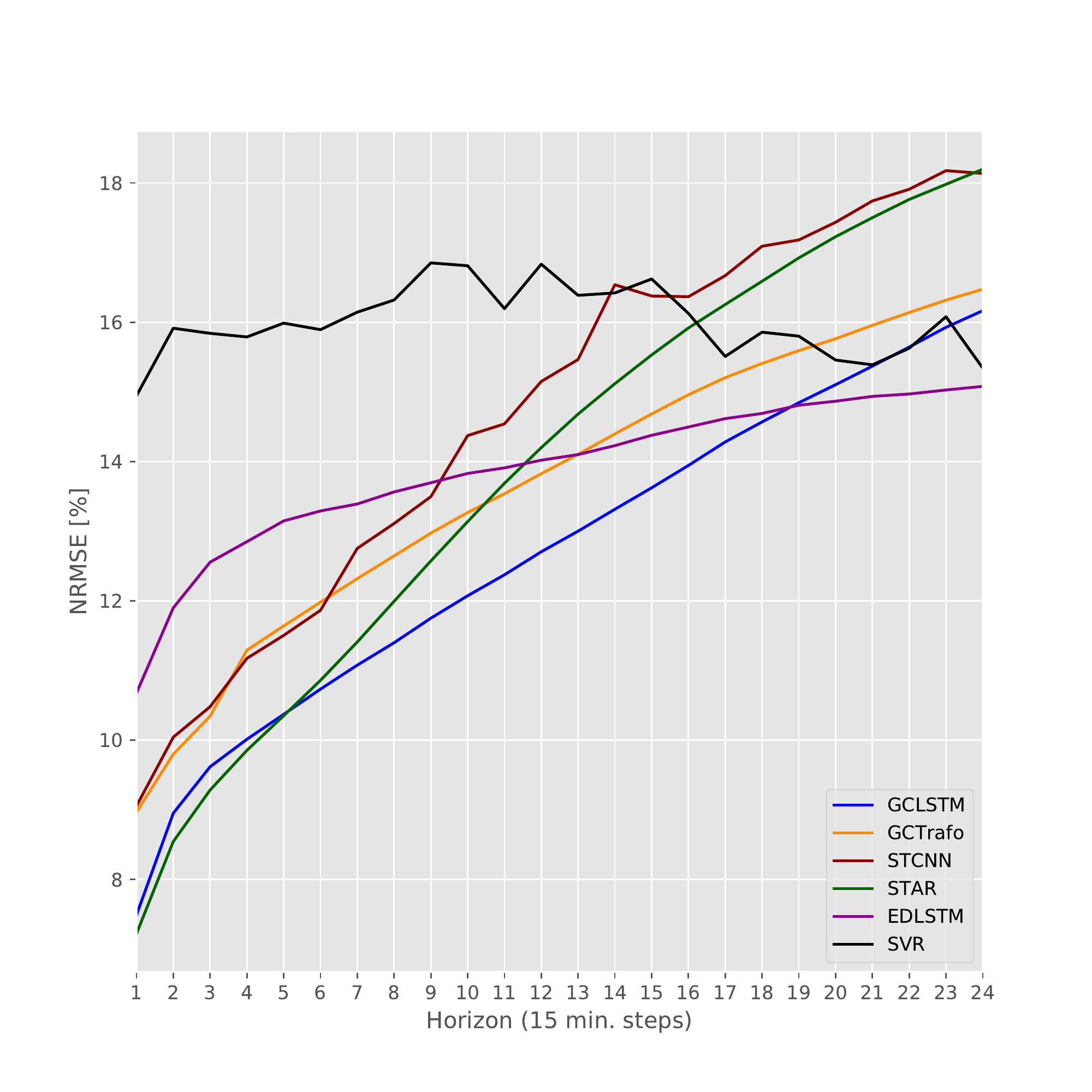}
     \caption{Forecast NRMSE for Bern.}
       \label{fig:nrmse_bern}
  \end{subfigure}
  \caption{Single-site error comparison between the proposed models (GCLSTM and GCTrafo), alternative multi-site methods with similar inputs (STAR and STCNN), and models that use NWP as inputs (SVR and EDLSTM).}
\label{fig:nrmse_single}
\end{figure*}

We analyzed the distance between nodes for which the model performances start to degrade on the real data set. To this end, the distance to the closest neighbor for each node is calculated and isolated nodes are found. Figure \ref{fig:knn1} shows the NMRSE for 1, 3 and 6 hours ahead predictions versus the distance to the closest neighbor for all nodes. The analysis does not indicate a higher error for sites that are further away and more isolated. For instance, the NRMSE for nodes with close neighbors (less than 5km) is between 10\% and 19\% for 3 hours ahead predictions. On the other hand, the NRMSE for the same horizon of isolated sites, i.e. 30 to 43 km away from the closest node, is between 11\% and 17\%. The same behavior is shown for all other prediction horizons. Therefore, up to 40km, the models don't show a drop in performance. 

Next, we show the forecasting results for two sites in the central part of Switzerland: Bern and Bätterkinden. The two locations are about 25 km apart. Figure \ref{fig:nrmse_single} shows the NRMSE evolution for GCLSTM, GCTrafo, STAR and STCNN, and the EDLSTM and SVR methods that use NWP as inputs. The error trend in  the multi-site comparison, between GCLSTM, GCTrafo, STAR and STCNN is similar to the one observed in the single-site comparison. For both sites, GCLTSM outperforms other methods between 1 and 4 hours ahead predictions. Interestingly, during the first hour (4 steps ahead) GCLSTM is on a par with the linear STAR method. However, for longer term forecasts, 5 to 6 hours ahead, EDLSTM and SVR methods yield lower errors than the proposed methods. The main reason lies in the fact that they use as the additional input NWP data, which has higher accuracy for six hours to day ahead predictions. However, NWP-based forecasts have higher error rate in comparison to other methods for intra-day forecasts. Additionally, we observe the high impact of the temporal resolution of the weather data  on the results, since a higher resolution of the weather data leads to higher accuracy of the forecast, which is the case of Bätterkinden (Figure \ref{fig:nrmse_batterkinden}). 

\begin{figure*}[h!]
  \centering
  \begin{subfigure}{0.5\linewidth}
    \includegraphics[width=\linewidth]{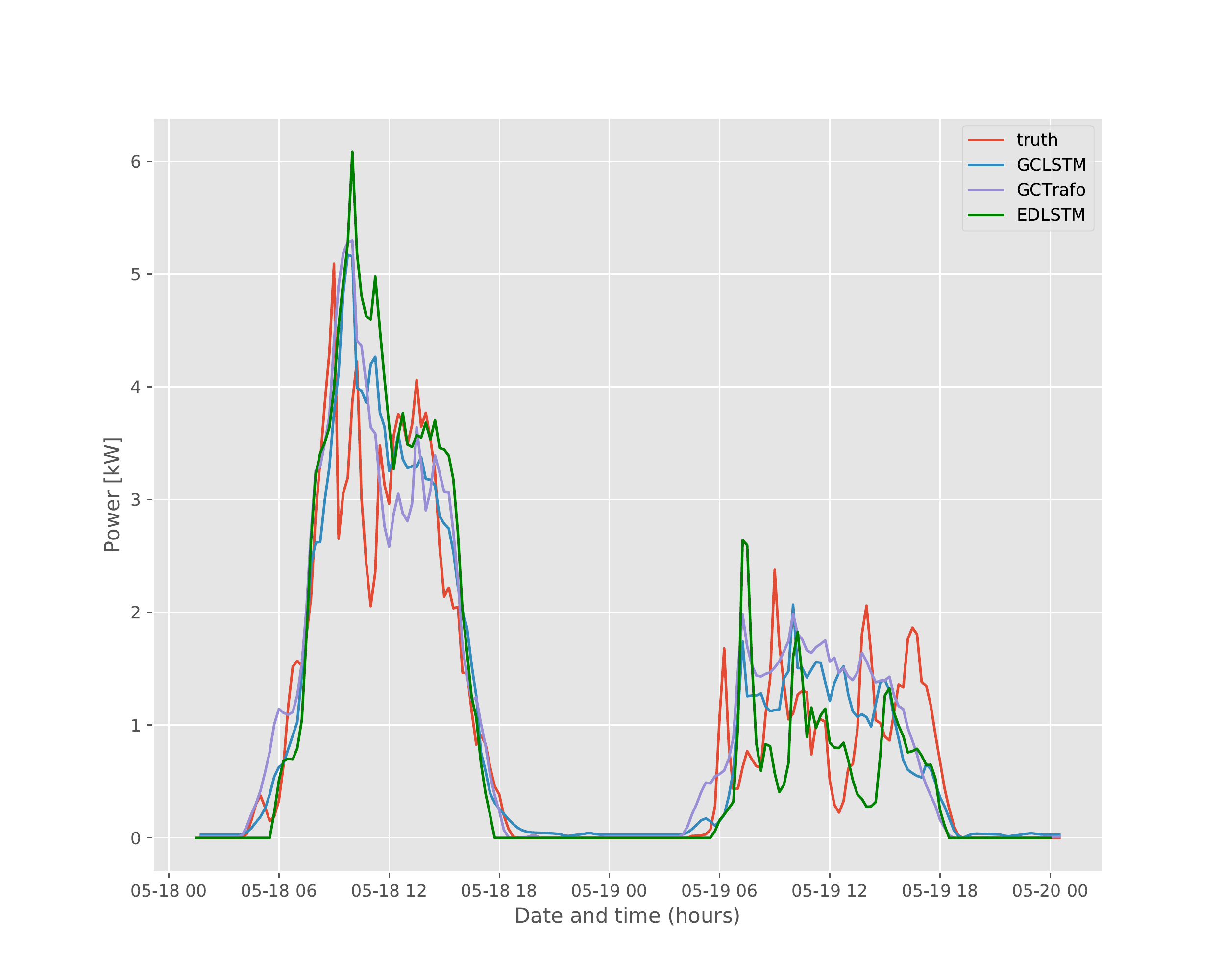}
    \caption{Measured production and 1 hour ahead predictions.}
      \label{fig:Bern_1}
  \end{subfigure}\hfill
  \begin{subfigure}{0.5\linewidth}
    \includegraphics[width=\linewidth]{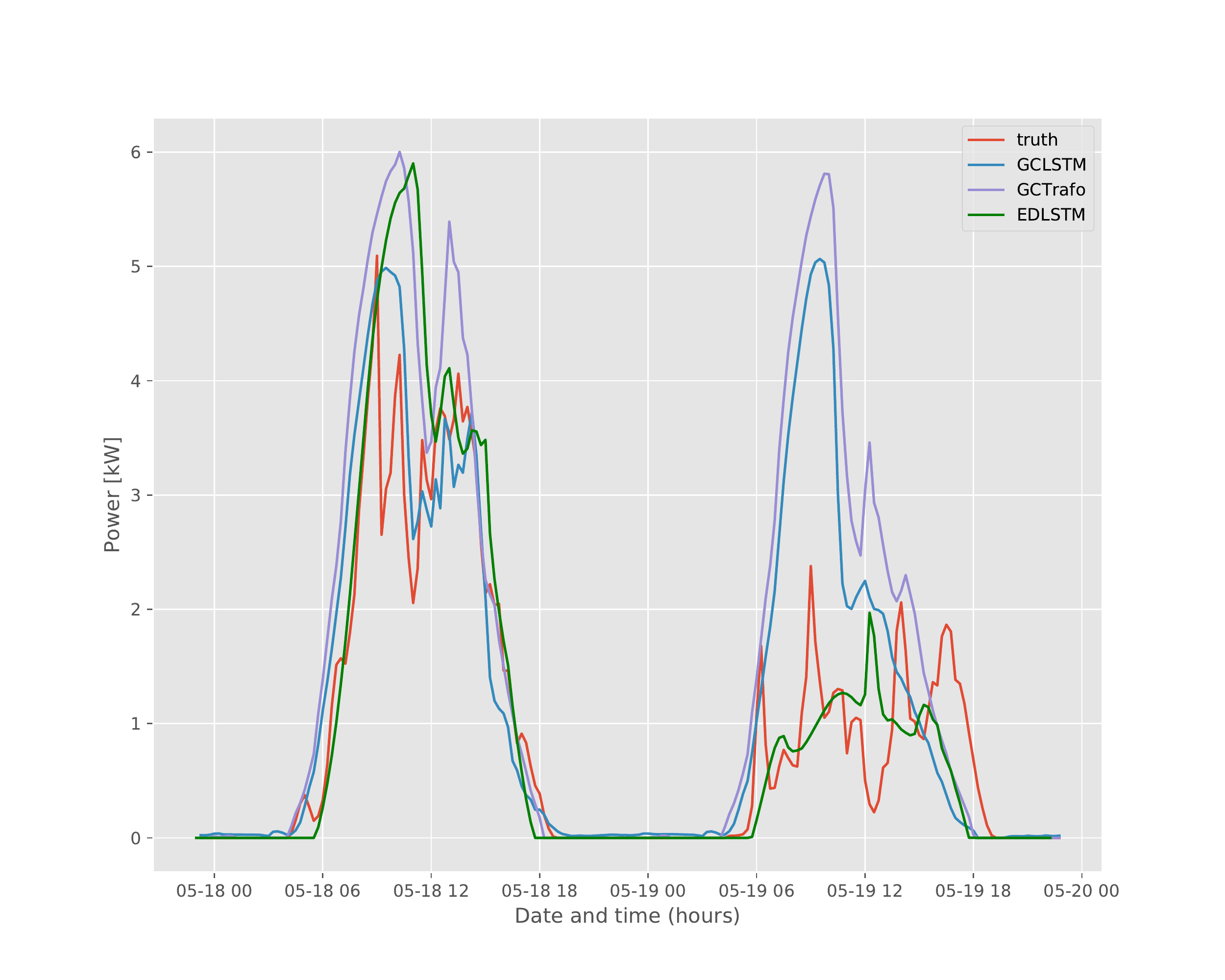}
    \caption{Measured production and 6 hour ahead predictions.}
      \label{fig:Bern_6}
  \end{subfigure}
  \caption{Illustration of measured and forecasted power production for two days in Bern. Only forecasts from GCLSTM, GCTrafo and EDLSTM are included.}
  \label{fig:Bern_1_6}
\end{figure*}
 
As an illustration of some of the advantages and limitations of the proposed methods, Figure \ref{fig:Bern_1_6} shows a visualization of the time series for one hour and six hours ahead forecasts for two days in Bern. The first day has a clear sky with a few clouds passing by during the middle of the day whereas the second day is a cloudy day with low production during whole day. The daily NRMSE for the two days and the two horizons are shown in Table \ref{tab:daily_nrmse}.  This visual comparison is made to show an extreme case where the proposed models might fail to provide an accurate forecast for long-term horizons (6h ahead), especially at the beginning of the day, and where models that use NWP as inputs (EDLSTM) have an advantage. 

Figure \ref{fig:Bern_1} shows time series of the true PV production and 1 hour ahead forecasts using GCLSTM, GCTrafo and EDLSTM. From the daily errors and the visual assessment, we can conclude that during cloudy days, for short-term forecasts, graph-based methods outperform EDLSTM because of their ability to capture cloud movement and spatial information. On the other hand EDLSTM relays on NWP that have low spatial and temporal resolution yielding poor forecasts, even though it uses past site data to initialize the encoder. 

\begin{table}[h!]
\caption{Daily NRMSE for Bern illustration}
\begin{center}
\begin{tabular}{|c|c|c|c|c|}
\hline 
\rule[-1ex]{0pt}{2.5ex} Model & Day1 1h & Day2 1h & Day1 6h & Day2 6h \\ 
\hline 
\rule[-1ex]{0pt}{2.5ex} GCLSTM & 8.79 & 7.96 & 8.94 & 20.87 \\ 
\hline 
\rule[-1ex]{0pt}{2.5ex} GCTrafo & 9.31 & 8.22 & 15.46 & 26.74 \\ 
\hline 
\rule[-1ex]{0pt}{2.5ex} EDLSTM & 11.85 & 8.91 & 13.98 & 7.71 \\ 
\hline 
\end{tabular}
\end{center}
\label{tab:daily_nrmse}  
\end{table}

However, for 6 hours ahead, GCLSTM and GCTrafo forecasts in Figure \ref{fig:Bern_6} show a bias during the first six hours of the day after sunrise. Since night PV production values are zero, the graph-based architectures only receive clear sky data and average power from the two previous days. Thus, GCTrafo and GCLSTM forecast higher production values during the first 6 hours after the sunrise. Once the graph-based architectures start to receive non-zero PV power information from the day, these methods start to correct the predictions and we observe a sudden drop in the predicted values. During very cloudy days, such as the second day, EDLSTM benefits from NWP for 6 hours ahead forecasts.

\section{Conclusions}
\label{section_5}
Two novel graph convolutional neural network architectures for multi-site  deterministic   PV generation forecasting, GCLSTM and GCTrafo, have been introduced and compared with state-of-the-art algorithms, both at single and multi-site level.
The extensive comparison on two PV power generation datasets (the real dataset with 304 plants and the synthetic dataset with 1000 PV plants)  has shown that they outperform state-of-the-art methods, with an average NRMSE error over the entire horizon (6 hours ahead) of  8.3\% (GCLSTM) and 8.4\% (GCTrafo) on the synthetic dataset, and 12.6\%  (GCLSTM) and 13.6\% (GCTrafo) on the real dataset. Both architectures were trained on a single GPU for the 1000 nodes case and can be scaled to a higher number of nodes using multi-GPU computing, making them appealing for grid management applications with large number of nodes. 

In forthcoming works, we will address some inherent limitations in the way spatio-temporal information is diffused across the nodes in these models. The number of nodes taken into account within the graph convolutions were limited to the $K$ closest neighbors because of the increase in computational complexity. However, it is expected that further away nodes might be important predictors if advection is dominant in the regional cloud dynamics at a specific time. Another research direction is to investigate the robustness and adaptability of the models to incomplete datasets. Many real world scenarios include datasets with missing data, as well as the addition of the new stations (nodes) without historical information which is crucial for the autoregressive models. Therefore, the robustness of the presented models will be studied and methods to adapt the models to unseen nodes will be investigated. Finally, another possible avenue of research is to transform the proposed deterministic models into probabilistic models by integrating noise into the deterministic model to build a generator in a similar fashion to \cite{koochali2020like}. This generator should be trained using an appropriate classifier as discriminator in an adversarial setting to make a probabilistic forecast. 

\appendix
In our experiments, the number of hidden dimensions $\rm{lat}$ in the encoder and decoder cells of the GCLSTM network (see Section \ref{sec:gclstm})  were equal to 32. The size of the MLP  at the end of the GCLSTM decoder was equal to [8, 48, 48]. For the GCTrafo, the following hyperparameters were chosen: the 1D-convolutional kernel was of size 4, the encoder and decoder convolutional latent spaces were of size  8, and 8 attention heads were used. The STCNN architecture had three 2D convolutional layers, with channel sizes of [128, 64, 32] and a kernel size of 11. Batch normalization and max pooling were applied after each convolutional layer. The single-site Encoder Decoder LSTM (EDLSTM) had a latent representation size of 64 and the decoder was followed by a MLP of size [64, 32]. STCNN, GCLSTM and GCTrafo models were trained with stochastic gradient descent and  Adam optimizer, without regularization. EDLSTM was trained with dropout as regularization. Additional hyperparameters are presented in Table \ref{tab:hyperparameters}. Finally, the STAR model used 3 hours of past time steps to forecast the PV production over the 6 hours horizon.

\begin{table*}[h!]
\caption{Table of hyperparameters}
\begin{center}
\begin{tabular}{clclccccc}
\hline
\multicolumn{2}{|l|}{Models} & \multicolumn{2}{c|}{\begin{tabular}[c]{@{}c@{}}Iterations\\  (real/synthetic)\end{tabular}} & \multicolumn{1}{c|}{\begin{tabular}[c]{@{}c@{}}Batch   size\\  (real/synthetic)\end{tabular}} & \multicolumn{1}{c|}{Past time steps - M} & \multicolumn{1}{c|}{\begin{tabular}[c]{@{}c@{}}k-nearest neighbours\\  (graph construction)\end{tabular}} & \multicolumn{1}{c|}{\begin{tabular}[c]{@{}c@{}}Order of Chebyshev \\ polynomial\end{tabular}} & \multicolumn{1}{c|}{\begin{tabular}[c]{@{}c@{}}Learning   rate / \\ Dropout rate\end{tabular}} \\ \hline
\multicolumn{2}{|c|}{GCTrafo}                 & \multicolumn{2}{c|}{70 000}                                                                 & \multicolumn{1}{c|}{64}                                                                       & \multicolumn{1}{c|}{16}                  & \multicolumn{1}{c|}{24}                                                                                  & \multicolumn{1}{c|}{2}                                                                        & \multicolumn{1}{c|}{1e-4 / -}                                                                \\ \hline
\multicolumn{2}{|c|}{GCLSTM}                  & \multicolumn{2}{c|}{50 000}                                                                 & \multicolumn{1}{c|}{64}                                                                       & \multicolumn{1}{c|}{16}                  & \multicolumn{1}{c|}{15}                                                                                  & \multicolumn{1}{c|}{4}                                                                        & \multicolumn{1}{c|}{1e-4 / -}                                                                  \\ \hline
\multicolumn{2}{|c|}{STCNN}                   & \multicolumn{2}{c|}{6000 / 10 000}                                          & \multicolumn{1}{c|}{128 /64}                                                                  & \multicolumn{1}{c|}{72}                  & \multicolumn{1}{c|}{-}                                                                                   & \multicolumn{1}{c|}{-}                                                                        & \multicolumn{1}{c|}{1e-4 / -}                                                                  \\ \hline
\multicolumn{2}{|c|}{EDLSTM}                  & \multicolumn{2}{c|}{30 000}                                                                 & \multicolumn{1}{c|}{128}                                                                      & \multicolumn{1}{c|}{16}                  & \multicolumn{1}{c|}{-}                                                                                   & \multicolumn{1}{c|}{-}                                                                        & \multicolumn{1}{c|}{1e-4 / 0.05}                                                               \\ \hline
\multicolumn{1}{l}{}            &             & \multicolumn{1}{l}{}                                   &                                    & \multicolumn{1}{l}{}                                                                          & \multicolumn{1}{l}{}                     & \multicolumn{1}{l}{}                                                                                     & \multicolumn{1}{l}{}                                                                          & \multicolumn{1}{l}{}    
\label{tab:hyperparameters}                                                                      
\end{tabular}
\end{center}
\end{table*}

\bibliography{PV_bibliography}
\bibliographystyle{IEEEtran}

\end{document}